%% file: arXiv.tex
\PassOptionsToPackage{table,xcdraw}{xcolor} 
\documentclass[10pt,twocolumn,letterpaper]{article}

\usepackage[pagenumbers]{cvpr} 

\input{preamble}

\definecolor{cvprblue}{rgb}{0.21,0.49,0.74}
\usepackage[pagebackref,breaklinks,colorlinks,citecolor=cvprblue]{hyperref}

\newcommand{\methodname}{{\tt{pFedES}}}
\usepackage[ruled]{algorithm2e}
\SetKwInput{KwInput}{Input}                
\SetKwInput{KwOutput}{Output}              
\definecolor{ylp_color1}{RGB}{143,187,222}
\definecolor{ylp_color2}{RGB}{190,215,248}
\usepackage{tcolorbox}
\usepackage{multirow}
\usepackage{booktabs}
\usepackage{wrapfig}
\usepackage{amsmath}
\usepackage{amsthm}
\usepackage{amsfonts}
\usepackage{bm}
\usepackage{eucal}
\usepackage{arydshln}

\newtheorem{theorem}{Theorem}

\newtheorem{lemma}{Lemma}

\newtheorem{assumption}{Assumption}

\usepackage{pifont}
\usepackage{setspace}
\usepackage{url}

\title{pFedES: Model Heterogeneous Personalized Federated Learning \\ with Feature Extractor Sharing}

\author{
Liping Yi$^{a}$, Han Yu$^{b,}$\thanks{Corresponding author}, Gang Wang$^{a,}$\thanks{Corresponding author}, Xiaoguang Liu$^{a}$\\
$^{a}~$College of Computer Science,
Nankai University,
Tianjin, China\\
$^{b~}$School of Computer Science and Engineering,
Nanyang Technological University,
Singapore\\
{\tt\small \{yiliping,wgzwp,liuxg\}@nbjl.nankai.edu.cn,han.yu@ntu.edu.sg}
}

\begin{document}
\maketitle
\input{sec/0_abstract}    
\input{sec/1_intro}

\input{sec/2_related}

\input{sec/3_preliminary}

\input{sec/4_method}
\input{sec/5_theorey}

\input{sec/6_experiment}
\input{sec/7_conclusion}

{   
    \small
    \bibliographystyle{ieeenat_fullname}
    \bibliography{main}
}

\input{sec/X_suppl}

\end{document}

%% file: preamble.tex
\usepackage[dvipsnames]{xcolor}


%% file: sec/0_abstract.tex
\begin{abstract}
As a privacy-preserving collaborative machine learning paradigm, federated learning (FL) has attracted significant interest from academia and the industry alike. 
To allow each data owner (a.k.a., FL clients) to train a heterogeneous and personalized local model based on its local data distribution, system resources and requirements on model structure, the field of model-heterogeneous personalized federated learning (MHPFL) has emerged. Existing MHPFL approaches either rely on the availability of a public dataset with special characteristics to facilitate knowledge transfer, incur high computation and communication costs, or face potential model leakage risks.
To address these limitations, we propose a model-heterogeneous \underline{p}ersonalized \underline{Fed}erated learning approach based on feature \underline{E}xtractor \underline{S}haring (\methodname{}). It incorporates a small homogeneous feature extractor into each client's heterogeneous local model. Clients train them via the proposed iterative learning method to enable the exchange of global generalized knowledge and local personalized knowledge. The small local homogeneous extractors produced after local training are uploaded to the FL server and for aggregation to facilitate easy knowledge sharing among clients.
We theoretically prove that \methodname{} can converge over wall-to-wall time.
Extensive experiments on two real-world datasets against six state-of-the-art methods demonstrate that \methodname{} builds the most accurate model, while incurring low communication and computation costs. Compared with the best-performing baseline, it achieves $1.61\%$ higher test accuracy, while reducing communication and computation costs by $99.6\%$ and $82.9\%$, respectively.
\end{abstract}

%% file: sec/1_intro.tex
\section{Introduction}
\label{sec:intro}

Federated learning (FL) \cite{Goebel-et-al:2023} is an emerging collaborative machine learning paradigm. It often relies on a central FL server to coordinate decentralized data owners (a.k.a., FL clients) to train a shared global FL model in a privacy-preserving manner \cite{1w-survey}.
Due to its potential to help artificial intelligence (AI) applications comply with privacy regulations, FL has been widely adopted in various fields, including computer vision (CV) \cite{Liu-et-al:2020FedVision}, healthcare \cite{Liu-et-al:2022IAAI} and power generation \cite{chen2023efficient}. In a traditional FL system, the server first broadcasts the global model to clients. Clients then train the received global model on their respective local dataset and upload the trained local models to the server. The server aggregates the received local models to update the global model. These steps are repeated until the global model converges. During the entire training process, only models are transmitted between the server and clients, while the data never leaves clients, thereby protecting data privacy.

The above prevailing mode of FL follows the model homogeneity assumption. It requires that all clients train local models with the same structure as the global model. Thus, it is still equipped to address the following important challenges often encountered in practice:
\begin{itemize}
    \item \textbf{Data Heterogeneity}. Clients can own non-independently and identically distributed (non-iid) data \cite{PFL-yu}. Directly aggregating biased Local models trained on such data can lead to sub-optimal global models \citep{Non-IID}. 
    
    \item \textbf{Resource Heterogeneity}. FL clients are often devices (e.g., mobile phones, autonomous vehicles) with divergent system resources in terms of computational power and communication bandwidth \citep{PruneFL}. Low-end devices can only train small models, while high-end devices can train large models. In the face of such resource heterogeneity, the traditional FL approach is only able to limit all clients to train the smallest denomination of the model, leading to model performance bottlenecks and wasted system resources for high-end devices. 
    
    \item \textbf{Model Ownership Heterogeneity}. In cases where FL clients are companies, they often fine-tune models from their internal repositories via FL. Different companies often maintain models with distinct structures. Due to intellectual property considerations \citep{MHPFL-survey}, they are reluctant to expose their models to others. Therefore, it is desirable to allow each FL client to train a local model following a unique structure to prevent model leakage to others.
\end{itemize}

To address the above challenges simultaneously, the field of model-heterogeneous personalized federated learning (MHPFL) has emerged. It focuses on enabling each FL client to train a personalized and heterogeneous model based on its local data distribution, system resources, and model structure requirements \citep{yi2023fedgh}. 
Prior efforts for MHPFL can be divided into three main branches: 1) knowledge distillation, 2) mutual learning, and 3) model mixup. 

\textit{Knowledge distillation-based MHPFL} methods either depend on a public dataset which is not always available \citep{FedDF}, or introduce heavy communication costs \citep{FedGEMS}, computation overheads \citep{FCCL}, and risks of privacy leakage \citep{Breaching-FedMD,FedProto}. \textit{Mutual learning-based MHPFL methods} \citep{FML,FedKD} train a local heterogeneous large model and a homogeneous small model on clients with a mutual learning approach and share the homogeneous small models for information fusion across different clients. Since there is no discussion about the relationship between the two models in model capacity or structure, model performance is limited, with extra computational costs incurred by clients. \textit{Model mixup-based MHPFL} methods \citep{LG-FedAvg,FedRep} split each client's local heterogeneous model into a homogeneous part and a heterogeneous part, and only share the homogeneous parts across clients. Only part of a complete local model (partial knowledge) is being shared, which leads to model performance bottleneck and leakage of the shared model structure.

To tackle the above limitations of existing MHPFL methods, we propose an efficient model-heterogeneous \underline{p}ersonalized \underline{Fed}erated learning framework based on small homogeneous feature \underline{E}xtractor \underline{S}haring (\methodname{}). 
It incorporates a small homogeneous feature extractor into each client's local heterogeneous model. Clients train these two models following the proposed iterative learning method to facilitate the exchange of globally generalized knowledge and locally personalized knowledge. The updated homogeneous feature extractors are then uploaded to the FL server for aggregation to facilitate knowledge sharing among heterogeneous local models.
Since only the small homogeneous feature extractors are transmitted between the server and clients, \methodname{} incurs low communication costs and protects the privacy of local data as well as local model intellectual properties. Since only one extra small homogeneous feature extractor is being trained by each client, the extra computational overhead introduced is low.

Through theoretical analysis, we show the non-convex convergence rate of \methodname{} and prove it converges over wall-to-wall time. 
Extensive experiments on two real-world datasets against six state-of-the-art methods demonstrate that \methodname{} builds the most accurate model, while incurring low communication and computation costs. Compared with the best-performing baseline, it achieves $1.61\%$ higher test accuracy, while reducing communication and computation costs by $99.6\%$ and $82.9\%$, respectively.




%% file: sec/2_related.tex
\section{Related Work}
Existing MHPFL methods have two families: a) \textit{partially model-heterogeneous}, clients hold different subnets of the global model, and heterogeneous subnets can be aggregated on the server, such as {\tt{FedRolex}} \citep{FedRolex}, {\tt{HeteroFL}} \citep{HeteroFL}, {\tt{FjORD}} \citep{FjORD}, {\tt{HFL}} \citep{HFL}, {\tt{Fed2}} \citep{Fed2}, {\tt{FedResCuE}} \citep{FedResCuE}. b) \textit{fully model-heterogeneous}, clients can hold models with completely different structures. These fully heterogeneous local models cannot be aggregated directly on the server. This branch of work involves the following three categories.


\textbf{Knowledge Distillation-based MHPFL}. For \textit{public dataset-dependent} knowledge distillation-based MHPFL methods (such as {\tt{Cronus}} \citep{Cronus}, {\tt{FedGEMS}} \citep{FedGEMS}, {\tt{Fed-ET}} \citep{Fed-ET}, {\tt{FSFL}} \citep{FSFL}, {\tt{FCCL}} \citep{FCCL}, {\tt{DS-FL}} \citep{DS-FL}, {\tt{FedMD}} \citep{FedMD}, {\tt{FedKT}} \citep{FedKT}, {\tt{FedDF}} \citep{FedDF}, {\tt{FedHeNN}} \citep{FedHeNN}, {\tt{FedAUX}} \citep{FEDAUX}, {\tt{CFD}} \citep{CFD}, {\tt{FedKEMF}} \citep{FedKEMF} and {\tt{KT-pFL}} \citep{KT-pFL}), the server aggregates the output logits of local heterogeneous models on a public dataset to construct the global logits. Clients then calculate the distance between the global logits and the local logits of their respective local heterogeneous models on the same public dataset as a distillation loss for training the local heterogeneous model. 
However, the public dataset is not always accessible and should have the same distribution as private data. Besides, computing and transmitting the logits for each public data sample incurs high extra computation and communication costs. Data privacy may be compromised by paired-logits inversion attacks \cite{Breaching-FedMD}. 

For other knowledge distillation-based MHPFL methods \textit{not dependent on a public dataset}, {\tt{FedZKT}} \citep{FedZKT} and {\tt{FedGen}} \citep{FedGen} introduce zero-shot knowledge distillation to FL. They generate a public dataset by training a generator, which is time-consuming. {\tt{HFD}} \citep{HFD1,HFD2}, {\tt{FedGKT}} \citep{FedGKT}, {\tt{FD}} \citep{FD} and {\tt{FedProto}} \citep{FedProto} allow each client to upload the (average) logits or representations of its local seen-class samples to the server. Then, the server aggregates logits or representations by classes, and the updated global class-logits or representations are sent back to clients and used to calculate the distillation loss with local logits. Clients under these methods are required to calculate distillation loss for each local data sample, incurring high computation costs. Besides, the uploaded logits or representations and the corresponding class information might compromise privacy.

\textbf{Mutual Learning-based MHPFL}. In {\tt{FML}} \citep{FML} and {\tt{FedKD}} \citep{FedKD}, each client owns a small homogeneous model and a large heterogeneous model trained via mutual learning. The trained small homogeneous models are aggregated by the server, i.e., the small homogeneous models serve as information media to facilitate knowledge transfer among local large heterogeneous models. However, they do not explore the relationship between the two models in model structure and parameter capacity, which negatively affects the final model performance and incurs high computation costs for training an extra homogeneous model for clients.

\textbf{Model Mixup-based MHPFL}. These methods split each client's local model into two parts: a feature extractor and a classifier, and only one part is shared. {\tt{FedMatch}} \citep{FedMatch}, {\tt{FedRep}} \citep{FedRep}, {\tt{FedBABU}} \citep{FedBABU} and {\tt{FedAlt/FedSim}} \citep{FedAlt/FedSim} share the homogeneous feature extractor to enhance model generalization while preserving the personalized local classifier. In contrast, {\tt{FedClassAvg}} \citep{FedClassAvg}, {\tt{LG-FedAvg}} \citep{LG-FedAvg} and {\tt{CHFL}} \citep{CHFL} share the homogeneous classifier to improve model classification while preserving the personalized local feature extractor. Since only part of the entire model is shared, the final local heterogeneous models face performance bottlenecks. Besides, they are also prone to leaking the structure of the shared part of the model.

In contrast, \methodname{} incorporates the global information carried by the global feature extractor as prior knowledge into the local heterogeneous model to enhance its generalization. As only the additional homogeneous feature extractors are exchanged between the server and the clients for knowledge transfer, while local heterogeneous models always stay with the clients, \methodname{} preserves model structure privacy while facilitating collaboration among the heterogeneous models from different FL clients.

%% file: sec/3_preliminary.tex
\section{Preliminaries}
{\tt{FedAvg}} \citep{FedAvg} is a typical FL algorithm. It assumes that a FL system consists of one central FL server and $N$ FL clients. In each communication round, the server randomly selects a fraction $C$ of $N$ clients (the selected client set is $S$, $|S|=C \cdot N=K$), and broadcasts the global model $\mathcal{F}(\omega)$ ($\mathcal{F}(\cdot)$ is model structure, $\omega$ are model parameters) to them. A client $k$ trains the received global model $\mathcal{F}(\omega)$ on its local dataset $D_k$ ($D_k \sim P_k$, $D_k$ obeys distribution $P_k$, i.e., local data from different clients are non-IID) to obtain an updated local model $\mathcal{F}(\omega_k)$ via gradient descent, i.e., $\omega_k \gets \omega-\eta \nabla \ell(\mathcal{F}(\boldsymbol{x}_i; \omega), y_i)$. $\ell(\mathcal{F}(\boldsymbol{x}_i; \omega), y_i)$ is the loss of the global model $\mathcal{F}(\omega)$ on the sample $(\boldsymbol{x}_i,y_i) \in D_k$. The updated local model $\mathcal{F}(\omega_k)$ is uploaded to the server. The server aggregates the received local models from $K$ clients via weighted averaging to update the global model, i.e., $\omega=\sum_{k=0}^{K-1} \frac{n_k}{n} \omega_k$ ($n_k=|D_k|$ is the data volume of client $k$, $n=\sum_{k=0}^{N-1} n_k$ is total data volume of all clients).

In short, the typical FL algorithm requires all clients to train local models with the same structures (\textbf{homogeneous}), and its training objective is to minimize the average loss of the global model $\mathcal{F}(\omega)$ on all client data, i.e.,
\begin{equation}
\min _{\omega \in \mathbb{R}^d} \sum_{k=0}^{K-1} \frac{n_k}{n} \mathcal{L}_k(\mathcal{F}(\omega);D_k),
\end{equation}
where the parameters of the global model, $\omega$, are $d$-dimensional real numbers. $\mathcal{L}_k(\mathcal{F}(\omega);D_k)$ is the average loss of the global model $\mathcal{F}(\omega)$ on client $k$'s local data $D_k$.

The objective of this paper is to study MHPFL in the context of supervised image classification tasks. We assume that all clients execute the same set of image classification tasks, and different clients hold heterogeneous local models with different structures, i.e., $\mathcal{F}_k(\omega_k)$ ($\mathcal{F}_k(\cdot)$ is the heterogeneous model structure, $\omega_k$ denotes personalized model parameters). \methodname{} aims to minimize the loss sum of all local heterogeneous personalized models, i.e.,
\begin{equation}
\min _{\omega_0, \ldots, \omega_{K-1} \in \mathbb{R}^{d_0, \ldots, d_{K-1}}} \sum_{k=0}^{K-1} \mathcal{L}_k(\mathcal{F}_k(\omega_k);D_k),
\end{equation}
where the parameters $\omega_0, \ldots, \omega_{K-1}$ of local heterogeneous models are $d_0, \ldots, d_{K-1}$-dimensional real numbers.

%% file: sec/4_method.tex
\section{The Proposed \methodname{} Approach}
\begin{figure}[!b]
\centering
\includegraphics[width=\linewidth]{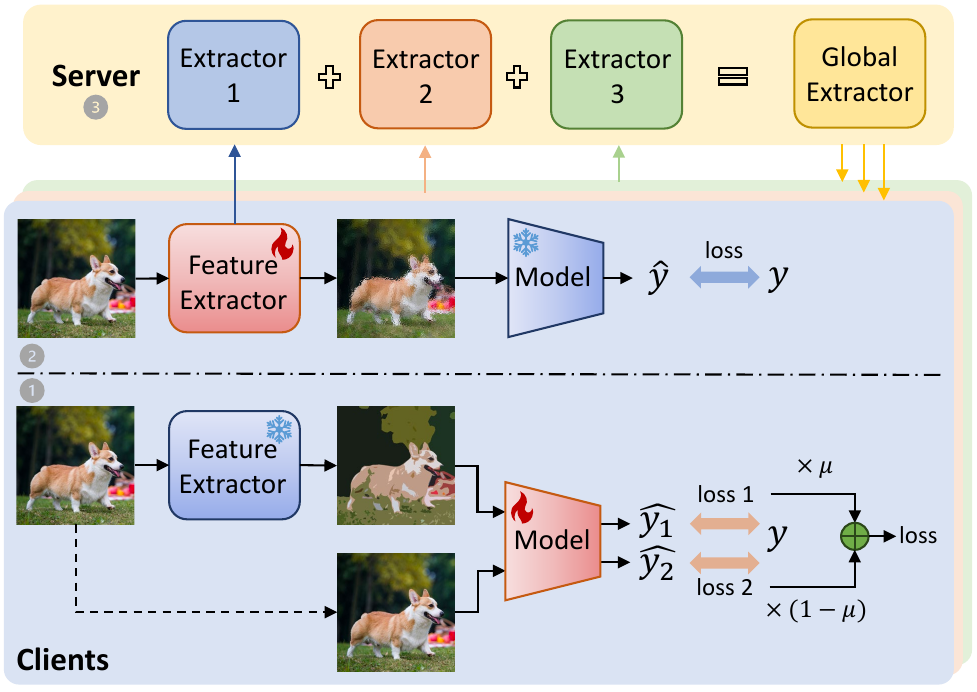}
\caption{Workflow of \methodname{}.}
\label{fig:FedES}
\end{figure}
To achieve the above design objective, \methodname{} incorporates an additional small homogeneous feature extractor $\mathcal{G}(\theta_k)$ ($\mathcal{G}(\cdot)$ is the structure of the homogeneous extractor, $\theta_k$ denotes the personalized parameters of client $k$'s extractor) into each FL client $k$. Clients then transfer knowledge across their heterogeneous local models by sharing their small homogeneous feature extractors.
In training round $t$, \methodname{} performs the following steps:
\begin{enumerate}
    \item The server selects $K$ clients among $N$ clients following some client selection approach \cite{Shi-et-al:2023}, and broadcasts the global feature extractor $\mathcal{G}(\theta^{t-1})$ aggregated in the $(t-1)$-th round to the selected clients.
    \item Client $k$ trains the received global feature extractor $\mathcal{G}(\theta^{t-1})$ and local model ${\mathcal{F}_k(\omega}_k^{t-1})$ on its local data $D_k$ following the proposed \textit{iterative training} method. Afterwards, the local feature extractor ${\mathcal{G}(\theta}_k^t)$ is uploaded to the FL server, while the heterogeneous local model ${\mathcal{F}_k(\omega}_k^t)$ remains within the client.
    \item The server aggregates the received client feature extractors ${\mathcal{G}(\theta}_k^t)$ to update the global feature extractor $\mathcal{G}(\theta^{t})$.
\end{enumerate}
The above steps are repeated until all clients' heterogeneous local models converge. Finally, each client's personalized heterogeneous local model is used for inference. The details of \methodname{} are described in Algorithm~\ref{alg:FedES}.

\begin{algorithm}[!t]
\small
\caption{\methodname{}}
\label{alg:FedES}
\KwInput{
$N$, total number of clients; $K$, number of selected clients in one round; $T$, total number of rounds; $\eta_\omega$, learning rate of heterogeneous local models; $\eta_\theta$, learning rate of local extractors; $\mu$, weight of loss for the combination of the frozen extractor and the training heterogeneous local model.
}
\KwOutput{$[\mathcal{F}_0(\omega_0^{T-1}),\ldots,\mathcal{F}_{N-1}(\omega_{N-1}^{T-1})]$.}
Randomly initialize heterogeneous local models $[\mathcal{F}_0(\omega_0^0),\ldots,\mathcal{F}_k(\omega_k^0),\ldots,\mathcal{F}_{N-1}(\omega_{N-1}^0)]$ and the global extractor $\mathcal{G}(\theta^0)$. \\ 
\For{each round t=1,...,T-1}{
    // \textbf{Server Side}: \\
    $S^t$ $\gets$ Randomly sample $K$ clients from $N$ clients; \\
    Broadcast the global extractor $\theta^{t-1}$ to $K$ clients; \\
    $\theta_k^t \gets$ \textbf{ClientUpdate}($\theta^{t-1}$); \\
    \begin{tcolorbox}[colback=ylp_color2,
                  colframe=ylp_color1,
                  width=4.3cm,
                  height=1cm,
                  arc=1mm, auto outer arc,
                  boxrule=1pt,
                  left=0pt,right=0pt,top=0pt,bottom=0pt,
                 ]
        /* Aggregate Local Extractor */ \\
        $\theta^t=\sum_{k=0}^{K-1}{\frac{n_k}{n}\theta_k^t}$.
    \end{tcolorbox}
    
    // \textbf{ClientUpdate}: \\
    Receive the global extractor $\theta^{t-1}$ from the server; \\
      \For{$k\in S^t$}{
    \begin{tcolorbox}[colback=ylp_color2,
                  colframe=ylp_color1,
                  width=5.9cm,
                  height=6.3cm,
                  arc=1mm, auto outer arc,
                  boxrule=1pt,
                  left=0pt,right=0pt,top=0pt,bottom=0pt,
                 ]
              /* Local Iterative Training */ \\
              // Freeze Extractor, Train Model \\
               \For{$(\boldsymbol{x},y)\in D_k$}{
                $\hat{\boldsymbol{x}}=\mathcal{G}(\theta^{t-1};\boldsymbol{x})$; \\
                $\widehat{y_1}={\mathcal{F}_k(\omega}_k^{t-1};\hat{\boldsymbol{x}});\widehat{y_2}={\mathcal{F}_k(\omega}_k^{t-1};\boldsymbol{x})$; \\
                $\ell_1=\ell(\widehat{y_1},y),\ \ell_2=\ell(\widehat{y_2},y)$; \\
                $\ell_\omega=\mu\cdot\ell_1+(1-\mu)\cdot\ell_2$; \\
                $\omega_k^t\gets\omega_k^{t-1}-\eta_\omega\nabla\ell_\omega$; 
               }
              // Freeze Model, Train Extractor \\
              \For{$(\boldsymbol{x},y)\in D_k$}{
                $\hat{\boldsymbol{x}}= \mathcal{G}(\theta^{t-1};\boldsymbol{x})$; \\
                $\hat{y}={\mathcal{F}_k(\omega}_k^t;\hat{\boldsymbol{x}})$; \\
                $\ell_\theta=\ \ell(\hat{y},y)$; \\
                $\theta_k^t\gets\theta^{t-1}-\eta_\theta\nabla\ell_\theta$;
              }
     \end{tcolorbox}
       Upload updated local extractor $\theta_k^t$ to the server.
    }
}
\textbf{Return} heterogeneous personalized local models $[\mathcal{F}_0(\omega_0^{T-1}),\mathcal{F}_1(\omega_1^{T-1}),\ldots,\mathcal{F}_k(\omega_k^{T-1}),\ldots,\mathcal{F}_{N-1}(\omega_{N-1}^{T-1})]$.  
\end{algorithm}

\subsection{Iterative Training} 
To facilitate effective global and local knowledge transfer, we design a novel iterative training method to train each client's homogeneous feature extractor and heterogeneous local model. 
\begin{enumerate}
    \item The global homogeneous feature extractor received from the server by a client is frozen. The local data are inputted as prompt information into the frozen global extractor carrying global knowledge to obtain the \textit{enhanced data} containing both global knowledge and local personalized knowledge. Then, both the local data and the enhanced data are used to train the heterogeneous local model to achieve \textit{global-to-local} knowledge transfer. 
    \item The heterogeneous local model is frozen after training in the first step. Then, the homogeneous feature extractor is trained with local data to achieve \textit{local-to-global knowledge} transfer. 
\end{enumerate}


\paragraph{Freeze Extractor, Train Local Model} Following Step \ding{192} as shown in Figure~\ref{fig:FedES}, in round $t$, for client $k$, freezing the global extractor $\mathcal{G}(\theta^{t-1})$ from the server and training
the heterogeneous local model ${\mathcal{F}_k(\omega}_k^{t-1})$ includes two data inputs: 1) client $k$'s original local data $(\boldsymbol{x},y)\in D_k$ being inputted into $\mathcal{G}(\theta^{t-1})$ to obtain the enhanced data $\hat{\boldsymbol{x}}=\mathcal{G}(\theta^{t-1};\boldsymbol{x})$ ($\hat{\boldsymbol{x}}$ and $\boldsymbol{x}$ are of the same dimension); 2) client $k$'s original local data $(\boldsymbol{x},y)\in D_k$. Both data are inputted into ${\mathcal{F}_k(\omega}_k^{t-1})$ to produce predictions:
\begin{equation}
    \widehat{y_1}={\mathcal{F}_k(\omega}_k^{t-1};\hat{\boldsymbol{x}});\widehat{y_2}={\mathcal{F}_k(\omega}_k^{t-1};\boldsymbol{x}).
\end{equation}
Then, $k$ calculates the hard loss (e.g., cross-entropy loss \cite{CEloss}) between the predictions $\widehat{y_1}$, $\widehat{y_2}$ and the label $y$,
\begin{equation}
    \ell_1=\ell(\widehat{y_1},y),\ \ell_2=\ell(\widehat{y_2},y).
\end{equation}

In earlier training rounds, the global extractor might be unstable. The enhanced data produced by a poor-performing global extractor might be of low quality, thereby negatively impacting the performance of the heterogeneous local model. To balance the global knowledge and the local personalized knowledge carried by the two types of input data, we compute a weighted sum of the hard loss of the two data inputs as the complete loss of the heterogeneous local model:
\begin{equation}\label{eq:miu}
    \ell_\omega=\mu\cdot\ell_1+(1-\mu)\cdot\ell_2,\ \mu\in(0,0.5].
\end{equation}
With the complete loss, the parameters of the heterogeneous local model are updated via gradient descent (e.g., SGD \cite{SGD}):
\begin{equation}
    \omega_k^t\gets\omega_k^{t-1}-\eta_\omega\nabla\ell_\omega.
\end{equation}
$\eta_\omega$ is the learning rate of the heterogeneous local model. During this step of training, the global knowledge from the frozen global extractor is transferred to the heterogeneous local model through the enhanced data, thereby improving its generalization. The local personalized knowledge in the original local data is further incorporated into the heterogeneous local model, thereby enhancing its personalization.

\paragraph{Freeze Local Model, Train Extractor} Following Step \ding{193} as shown in Figure~\ref{fig:FedES}, the heterogeneous local model ${\mathcal{F}_k(\omega}_k^t)$ trained in Step \ding{192} is frozen and the global feature extractor $\mathcal{G}(\theta^{t-1})$ is trained. Client $k$ inputs its local data $(\boldsymbol{x},y)\in D_k$ into $\mathcal{G}(\theta^{t-1})$ to generate the enhanced data $\hat{\boldsymbol{x}}= \mathcal{G}(\theta^{t-1};\boldsymbol{x})$. Then, it inputs $\hat{\boldsymbol{x}}$ into the frozen ${\mathcal{F}_k(\omega}_k^t)$ to obtain:
\begin{equation}
    \hat{y}={\mathcal{F}_k(\omega}_k^t;\hat{\boldsymbol{x}}).
\end{equation}
Then, $k$ computes the hard loss between the prediction $\hat{y}$ and label $y$ as:
\begin{equation}
    \ell_\theta=\ \ell(\hat{y},y).
\end{equation}
After obtaining the loss, $k$ updates the parameters of the feature extractor via gradient descent:
\begin{equation}
    \theta_k^t\gets\theta^{t-1}-\eta_\theta\nabla\ell_\theta.
\end{equation}
$\eta_\theta$ is the learning rate of the feature extractor. During this step, personalized local knowledge is transferred into the updated homogeneous local feature extractor, which in turn, is uploaded to the FL server for aggregation.

\subsection{Homogeneous Extractor Aggregation}
After receiving the local feature extractors ${\mathcal{G}(\theta}_k^t)$ from selected clients, the server aggregates them following {\tt{FedAvg}} to facilitate knowledge fusion across clients:
\begin{equation}
    \theta^t=\sum_{k=0}^{K-1}{\frac{n_k}{n}\theta_k^t}.
\end{equation}

\subsection{Feature Extractor Structure}
\begin{figure}[t]
\centering
\includegraphics[width=1\linewidth]{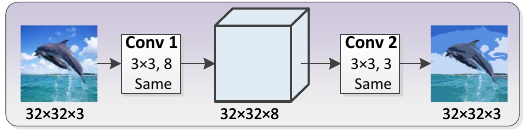}
\caption{The structure of the homogeneous extractor in \methodname{}.}
\label{fig:extractor}
\end{figure}

Since FL clients often are mobile edge devices with limited computational power, \methodname{} must improve the performance of local heterogeneous locals without introducing too much computation costs. Therefore, we design a small CNN model consisting of two convolutional layers with \textit{`padding=same'} as the feature extractor, as shown in Figure~\ref{fig:extractor}. Setting `padding=same' is to guarantee the dimensions of the original data input into the feature extractor and the enhanced data output by the feature extractor are the same. We use this small CNN as the default feature extractor in subsequent experiments.

\subsection{Discussion}
Here, we discuss \methodname{}'s computational costs, communication overheads and privacy preservation.

\textbf{Computational Cost}. On top of training heterogeneous local models, clients also train an additional small homogeneous feature extractor. Since we design a small CNN with two convolutional layers as the feature extractor, training it introduces low computational costs.

\textbf{Communication Overhead}. Since only the small homogeneous feature extractors are transmitted between the server and the clients, \methodname{} incurs much lower communication overhead than transmitting the complete models as in {\tt{FedAvg}}.

\textbf{Privacy Preservation}. Since only the parameters of the homogeneous feature extractors are transmitted between the server and the clients while local data and local models never leave the clients, \methodname{} preserves data privacy and model intellectual property.

%% file: sec/5_theorey.tex
\section{Analysis}
Before analyzing the convergence of \methodname{}, we first declare some additional notations. We use $t$ to denote a communication round and $e \in \{0,1,\ldots,E\}$ to denote the epochs/iterations of local training. In each round, each client executes $E$ local training iterations. $tE+e$ is the $e$-th iteration in the $(t+1)$-th round. $tE+0$ indicates that in the $(t+1)$-th round, before local model training, clients receive the global extractor $\mathcal{G}(\theta^t)$ aggregated in the $t$-th round. $tE+E$ is the last iteration of local training, indicating the end of local training in the $(t+1)$-th round. We denote the combination of the frozen homogeneous feature extractor with $\mathcal{G}(\theta)$, and the training of the heterogeneous local model $\mathcal{F}_k(\omega_k)$ at the first branch of Step 1 during the iterative training process as $\mathcal{H}_k(\varphi_k) = \mathcal{G}(\theta) \circ \mathcal{F}_k(\omega_k)$. We also assume that$\mathcal{F}_k(\omega_k)$ and the combined model $\mathcal{H}_k(\varphi_k)$ use the same learning rate $\eta=\eta_\omega=\eta_\varphi$.

\begin{assumption}\label{assump:Lipschitz}
\textbf{Lipschitz Smoothness}. The gradients of Client $k$'s heterogeneous local model are $L1$--Lipschitz smooth, i.e.,
\begin{equation}\label{eq:7}
\footnotesize
\begin{gathered}
\|\nabla \mathcal{L}_k^{t_1}(\omega_k^{t_1} ; \boldsymbol{x}, y)-\nabla \mathcal{L}_k^{t_2}(\omega_k^{t_2} ; \boldsymbol{x}, y)\| \leqslant L_1\|\omega_k^{t_1}-\omega_k^{t_2}\|, \\
\forall t_1, t_2>0, k \in\{0,1, \ldots, N-1\},(\boldsymbol{x}, y) \in D_k.
\end{gathered}
\end{equation}
The above formulation can be expressed as:
\begin{equation}
\footnotesize
\mathcal{L}_k^{t_1}-\mathcal{L}_k^{t_2} \leqslant\langle\nabla \mathcal{L}_k^{t_2},(\omega_k^{t_1}-\omega_k^{t_2})\rangle+\frac{L_1}{2}\|\omega_k^{t_1}-\omega_k^{t_2}\|_2^2 .
\end{equation}
\end{assumption}

\begin{assumption} \label{assump:Unbiased}
\textbf{Unbiased Gradient and Bounded Variance}. The random gradient $g_{\omega,k}^t=\nabla \mathcal{L}_k^t(\omega_k^t; \mathcal{B}_k^t)$ ($\mathcal{B}$ is a batch of local data) of each client's heterogeneous local model $\mathcal{F}_k(\omega_k)$ is unbiased, and the random gradient $g_{\varphi,k}^t=\nabla\mathcal{L}_k^t(\varphi_k^t;\mathcal{B}_k^t)$ of each client's combined model $\mathcal{H}_k(\varphi_k)$ is also unbiased, i.e.,
\begin{equation}
\footnotesize
\begin{split}
\mathbb{E}_{\mathcal{B}_k^t \subseteq D_k}[g_{\omega,k}^t]=\nabla \mathcal{L}_k^t(\omega_k^t), \\
\mathbb{E}_{\mathcal{B}_k^t\subseteq D_k}[g_{\varphi,k}^t]=\nabla\mathcal{L}_k^t(\varphi_k^t),
\end{split}
\end{equation}
and the variance of random gradient $g_{\omega,k}^t$ and $g_{\varphi,k}^t$ is bounded by:
\begin{equation}
\footnotesize
\begin{split}
\mathbb{E}_{\mathcal{B}_k^t \subseteq D_k}[\|\nabla \mathcal{L}_k^t(\omega_k^t ; \mathcal{B}_k^t)-\nabla \mathcal{L}_k^t(\omega_k^t)\|_2^2] \leqslant \sigma^2, \\
\mathbb{E}_{\mathcal{B}_k^t \subseteq D_k}[\|\nabla \mathcal{L}_k^t(\varphi_k^t ; \mathcal{B}_k^t)-\nabla \mathcal{L}_k^t(\varphi_k^t)\|_2^2] \leqslant \delta^2.
\end{split}
\end{equation}    
\end{assumption}

With these assumptions, we derive the following lemma and theorem (proofs can be found in Appendices~\ref{sec:proof-lemma} and \ref{sec:proof-theorem}).

\begin{lemma} \label{lemma:LocalTraining}
Based on Assumptions~\ref{assump:Lipschitz} and \ref{assump:Unbiased}, during local iterations $\{0,1,...,E\}$ in the $(t+1)$-th FL training round, the loss of an arbitrary client's heterogeneous local model is bounded by:
\begin{equation}
\footnotesize
\begin{aligned}
\mathbb{E}[\mathcal{L}_{(t+1) E}] &\leq \mathcal{L}_{t E+0}+(\frac{L_1 \eta^2 \tilde{\mu}^2}{2}-\eta \tilde{\mu}) \sum_{e=0}^{E-1}\|\nabla \mathcal{L}_{t E+e}\|_2^2 \\
&+\frac{L_1 \eta^2(\sigma^2+\delta^2)}{2}.
\end{aligned}
\end{equation}
where $\tilde{\mu} = 1- \mu$, $\mu \in (0,0.5]$, $\tilde{\mu} \in [0.5,1)$.
\end{lemma}

\begin{theorem} \label{theorem:non-convex}
\textbf{Non-convex convergence rate of \methodname{}}. Based on the above assumptions and lemma, for an arbitrary client and any $\epsilon>0$, the following inequality holds:
\begin{equation}
\footnotesize
\begin{aligned}
\frac{1}{T} \sum_{t=0}^{T-1} \sum_{e=0}^{E-1}\|\nabla \mathcal{L}_{t E+e}\|_2^2 & \leq \frac{\frac{1}{T} \sum_{t=0}^{T-1}(\mathcal{L}_{t E+0}-\mathbb{E}[\mathcal{L}_{(t+1) E}])}{\eta \tilde{\mu}-\frac{L_1 \eta^2 \tilde{\mu}^2}{2}} \\
& + \frac{\frac{L_1 \eta^2(\sigma^2+\delta^2)}{2}}{\eta \tilde{\mu}-\frac{L_1 \eta^2 \tilde{\mu}^2}{2}}
 <\epsilon, \\
s.t. \eta & <\frac{2 \epsilon \tilde{\mu}}{L_1(\sigma^2+\delta^2+\tilde{\mu}^2 \epsilon)}.
\end{aligned}
\end{equation}
\end{theorem}
Hence, under \methodname{}, a client's heterogeneous local model converges at a non-convex rate of $\epsilon \sim \mathcal{O}(\frac{1}{T})$.

%% file: sec/6_experiment.tex
\section{Experimental Evaluation}
To evaluate the performance of \methodname{}, we compared it against $6$ state-of-the-art MHPFL approaches on $2$ real-world datasets. The experiments are conducted with Pytorch on $4$ NVIDIA GeForce RTX 3090 GPUs with 24G memory.\footnote{Link to code omitted for double-blind review.}

\subsection{Experiment Setup}
\textbf{Datasets}. We evaluate \methodname{} and baselines on two image classification datasets: CIFAR-10 and CIFAR-100 \footnote{\url{https://www.cs.toronto.edu/\%7Ekriz/cifar.html}} \cite{cifar}. They are divided into non-IID datasets following the method specified in \cite{pFedHN} to facilitate MHPFL experimentation. For CIFAR-10, we assign only data from 2 out of the 10 classes to each client (non-IID: 2/10). For CIFAR-100, we assign only data from 10 out of the 100 classes to each client (non-IID: 10/100). Then, each client's local data are further divided into the training set, the validation set, and the test set following the ratio of 8:1:1. The test set is stored by each client and follows the same distribution as the local training set.

\textbf{Models}. As shown in Table~\ref{tab:model-structures}, each client trains a CNN model with output layer (i.e., the last fully connected layer) dimensions of $10$ and $100$ for CIFAR-10 and CIFAR-100, respectively. In model-homogeneous settings, each client has the same CNN-1 model structure. In model-heterogeneous settings, clients are randomly assigned with \{CNN-1,$\cdots$, CNN-5\} with uniform probability.
For both model-homogeneous and model-heterogeneous settings, the structure of the homogeneous feature extractor is as shown in Figure~\ref{fig:extractor}.

\begin{table}[t]
\centering
\resizebox{0.9\linewidth}{!}{%
\begin{tabular}{|l|c|c|c|c|c|}
\hline
Layer Name         & CNN-1    & CNN-2   & CNN-3   & CNN-4   & CNN-5   \\ \hline
Conv1              & 5$\times$5, 16   & 5$\times$5, 16  & 5$\times$5, 16  & 5$\times$5, 16  & 5$\times$5, 16  \\
Maxpool1              & 2$\times$2   & 2$\times$2  & 2$\times$2  & 2$\times$2  & 2$\times$2  \\
Conv2              & 5$\times$5, 32   & 5$\times$5, 16  & 5$\times$5, 32  & 5$\times$5, 32  & 5$\times$5, 32  \\
Maxpool2              & 2$\times$2   & 2$\times$2  & 2$\times$2  & 2$\times$2  & 2$\times$2  \\
FC1                & 2000     & 2000    & 1000    & 800     & 500     \\
FC2                & 500      & 500     & 500     & 500     & 500     \\
FC3                & 10/100   & 10/100  & 10/100  & 10/100  & 10/100  \\ \hline
model size & 10.00 MB & 6.92 MB & 5.04 MB & 3.81 MB & 2.55 MB \\ \hline
\end{tabular}%
}
\vspace{-0.5em}
\caption{Structures of $5$ heterogeneous CNN models with convolutional layers having $5 \times 5$ kernel size and $16$ or $32$ filters.}
\label{tab:model-structures}
\vspace{-1em}
\end{table}

\textbf{Baselines}. We compare \methodname{} with 6 baselines. 
\begin{itemize}
    \item {\tt{Standalone}}, clients train local models independently;
    \item Public-data independent knowledge distillation-based MHPFL methods: {\tt{FD}} \cite{FD} and {\tt{FedProto}} \cite{FedProto}; 
    \item Mutual learning-based MHPFL methods {\tt{FML}} \cite{FML} and {\tt{FedKD}} \cite{FedKD};
    \item Model mixup-based MHPFL methods {\tt{LG-FedAvg}} \cite{LG-FedAvg}.
\end{itemize}

\textbf{Evaluation Metrics}. \textbf{1) Accuracy}: we measure the \textit{individual test accuracy} of each client's heterogeneous local model and calculate the \textit{average test accuracy} of all clients' local models. \textbf{2) Communication Cost}: We trace the number of transmitted parameters when the average model accuracy reaches the target. \textbf{3) Computation Cost}: We track the computation FLOPs consumed when the average model accuracy reaches the target.

\textbf{Training Strategy}. We tune the optimal FL settings for all methods via grid search. The epochs of local model training $E \in \{1, 10\}$ and the batch size of local training $B \in \{64, 128, 256, 512\}$. The optimizer for local training is SGD with learning rate $\eta=\eta_\omega=\eta_\theta=0.01$. We also tune special hyperparameters for the baselines and report the optimal results. 
We adjust two hyperparameters ($\mu$ and $E_{fe}$, the loss weight and training epoch of the feature extractor) for \methodname{} to achieve the best performance. The detailed results of hyperparameter sensitivity tests are described in Appendix.~\ref{app:hyperparameter}. To compare with the baselines fairly, we set the total number of rounds $T \in \{100, 500\}$ to ensure that all algorithms converge.

\begin{table}[t]
\centering
\resizebox{\linewidth}{!}{%
\begin{tabular}{|l|cc|cc|cc|}
\hline
           & \multicolumn{2}{c|}{N=10, C=100\%}                                                                   & \multicolumn{2}{c|}{N=50, C=20\%}                                                                                           & \multicolumn{2}{c|}{N=100, C=10\%}                                                                  \\ \cline{2-7} 
Method     & \multicolumn{1}{c|}{CIFAR-10}                               & CIFAR-100                              & \multicolumn{1}{c|}{CIFAR-10}                               & CIFAR-100                                                     & \multicolumn{1}{c|}{CIFAR-10}                              & CIFAR-100                              \\ \hline
Standalone & \multicolumn{1}{c|}{96.35}                                  & \cellcolor[HTML]{D3D3D3}74.32                                  & \multicolumn{1}{c|}{\cellcolor[HTML]{D3D3D3}95.25}          & 62.38                                                         & \multicolumn{1}{c|}{\cellcolor[HTML]{D3D3D3}92.58}         & \cellcolor[HTML]{D3D3D3}54.93          \\
FML~\cite{FML}       & \multicolumn{1}{c|}{94.83}                                  & 70.02                                  & \multicolumn{1}{c|}{93.18}                                  & 57.56                                                         & \multicolumn{1}{c|}{87.93}                                 & 46.20                                  \\
FedKD~\cite{FedKD}      & \multicolumn{1}{c|}{94.77}                                  & 70.04                                  & \multicolumn{1}{c|}{92.93}                                  & 57.56                                                         & \multicolumn{1}{c|}{90.23}                                 & 50.99                                  \\
LG-FedAvg~\cite{LG-FedAvg}  & \multicolumn{1}{c|}{\cellcolor[HTML]{D3D3D3}96.47}          & 73.43                                  & \multicolumn{1}{c|}{94.20}                                  & 61.77                                                         & \multicolumn{1}{c|}{90.25}                                 & 46.64                                  \\
FD~\cite{FD}         & \multicolumn{1}{c|}{96.30}                                  & -                                      & \multicolumn{1}{c|}{-}                                      & -                                                             & \multicolumn{1}{c|}{-}                                     & -                                      \\
FedProto~\cite{FedProto}   & \multicolumn{1}{c|}{95.83}                                  & 72.79                                  & \multicolumn{1}{c|}{95.10}                                  & \cellcolor[HTML]{D3D3D3}62.55                                 & \multicolumn{1}{c|}{91.19}                                 & 54.01                                  \\ \hline
pFedES    & \multicolumn{1}{c|}{\cellcolor[HTML]{9B9B9B}\textbf{96.68}} & \cellcolor[HTML]{9B9B9B}\textbf{74.42} & \multicolumn{1}{c|}{\cellcolor[HTML]{9B9B9B}\textbf{95.74}} & \cellcolor[HTML]{9B9B9B}{\color[HTML]{000000} \textbf{63.55}} & \multicolumn{1}{c|}{\cellcolor[HTML]{9B9B9B}\textbf{92.89}} & \cellcolor[HTML]{9B9B9B}\textbf{55.15} \\ \hline
\end{tabular}%
}
\vspace{-0.5em}
\caption{Average accuracy ($\%$) for \textit{model-homogeneous} scenarios. $N$ is the total number of clients. $C$ is the fraction of participating clients in each round. `-' denotes failure to converge.}
\label{tab:compare-homo}
\vspace{-1em}
\end{table}

\begin{table}[t]
\centering
\resizebox{\linewidth}{!}{%
\begin{tabular}{|l|cc|cc|cc|}
\hline
           & \multicolumn{2}{c|}{N=10, C=100\%}                                                                & \multicolumn{2}{c|}{N=50, C=20\%}                                                                    & \multicolumn{2}{c|}{N=100, C=10\%}                                                                   \\ \cline{2-7} 
Method     & \multicolumn{1}{c|}{CIFAR-10}                            & CIFAR-100                              & \multicolumn{1}{c|}{CIFAR-10}                               & CIFAR-100                              & \multicolumn{1}{c|}{CIFAR-10}                               & CIFAR-100                              \\ \hline
Standalone & \multicolumn{1}{c|}{\cellcolor[HTML]{D3D3D3}96.53}       & 72.53                                  & \multicolumn{1}{c|}{95.14}                                  & \cellcolor[HTML]{D3D3D3}62.71          & \multicolumn{1}{c|}{91.97}                                  & 53.04                                  \\
FML~\cite{FML}        & \multicolumn{1}{c|}{30.48}                               & 16.84                                  & \multicolumn{1}{c|}{-}                                      & 21.96                                  & \multicolumn{1}{c|}{-}                                      & 15.21                                  \\
FedKD~\cite{FedKD}      & \multicolumn{1}{c|}{80.20}                               & 53.23                                  & \multicolumn{1}{c|}{77.37}                                  & 44.27                                  & \multicolumn{1}{c|}{73.21}                                  & 37.21                                  \\
LG-FedAvg~\cite{LG-FedAvg}  & \multicolumn{1}{c|}{96.30}                               & 72.20                                  & \multicolumn{1}{c|}{94.83}                                  & 60.95                                  & \multicolumn{1}{c|}{91.27}                                  & 45.83                                  \\
FD~\cite{FD}          & \multicolumn{1}{c|}{96.21}                               & -                                      & \multicolumn{1}{c|}{-}                                      & -                                      & \multicolumn{1}{c|}{-}                                      & -                                      \\
FedProto~\cite{FedProto}   & \multicolumn{1}{c|}{96.51}                               & \cellcolor[HTML]{D3D3D3}72.59          & \multicolumn{1}{c|}{\cellcolor[HTML]{D3D3D3}95.48}          & 62.69                                  & \multicolumn{1}{c|}{\cellcolor[HTML]{D3D3D3}92.49}          & \cellcolor[HTML]{D3D3D3}53.67          \\ \hline
pFedES    & \multicolumn{1}{c|}{\cellcolor[HTML]{9B9B9B}\textbf{96.70}} & \cellcolor[HTML]{9B9B9B}\textbf{73.89} & \multicolumn{1}{c|}{\cellcolor[HTML]{9B9B9B}\textbf{95.79}} & \cellcolor[HTML]{9B9B9B}\textbf{64.32} & \multicolumn{1}{c|}{\cellcolor[HTML]{9B9B9B}\textbf{92.72}} & \cellcolor[HTML]{9B9B9B}\textbf{54.40} \\ \hline
\end{tabular}%
}
\vspace{-0.5em}
\caption{Average accuracy for \textit{model-heterogeneous} scenarios.}
\label{tab:compare-hetero}
\vspace{-1em}
\end{table}

\subsection{Comparisons Results}
We compare \methodname{} with baselines under both model-homogeneous and model-heterogeneous scenarios with different $N$ and $C$ settings. We set up three scenarios: $\{(N=10, C=100\%), (N=50, C=20\%), (N=100, C=10\%)\}$. For ease of comparison across the three settings, $N\times C$ is set to be the same (i.e., $10$ clients participate in each round of FL). For {\tt{FML}} and {\tt{FedKD}}, under model-heterogeneous settings, we adopt the smallest `CNN-5' model as the small homogeneous model. 

\begin{figure}[t]
\centering
\begin{minipage}[t]{0.1666\textwidth}
\centering
\includegraphics[width=1.24in]{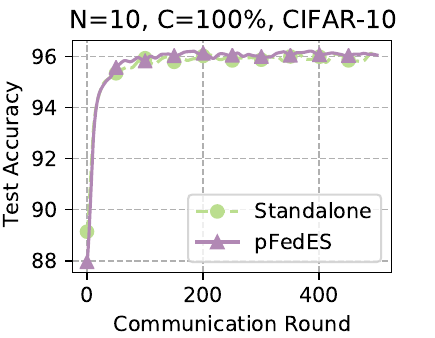}
\end{minipage}%
\begin{minipage}[t]{0.1666\textwidth}
\centering
\includegraphics[width=1.24in]{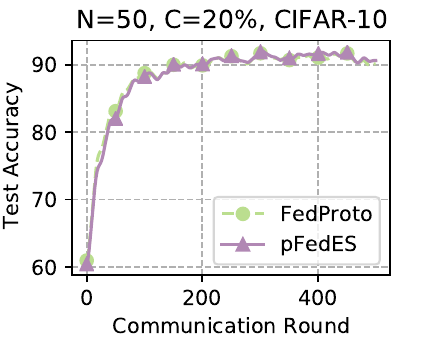}
\end{minipage}%
\begin{minipage}[t]{0.1666\textwidth}
\centering
\includegraphics[width=1.24in]{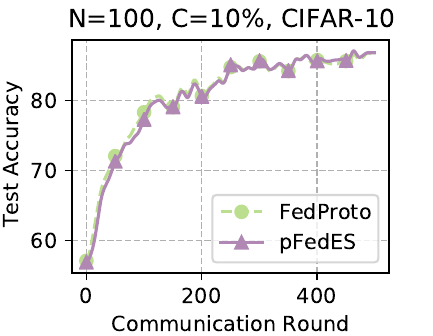}
\end{minipage}%

\begin{minipage}[t]{0.1666\textwidth}
\centering
\includegraphics[width=1.24in]{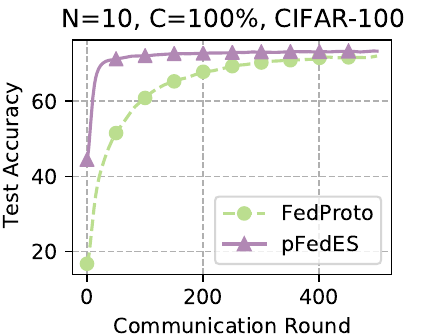}
\end{minipage}%
\begin{minipage}[t]{0.1666\textwidth}
\centering
\includegraphics[width=1.24in]{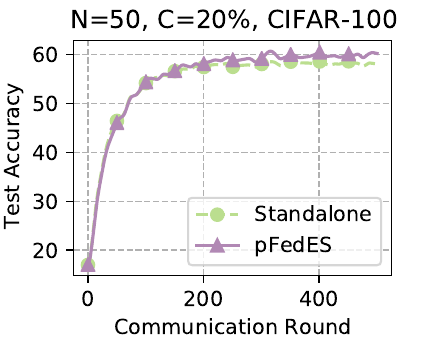}
\end{minipage}%
\begin{minipage}[t]{0.1666\textwidth}
\centering
\includegraphics[width=1.24in]{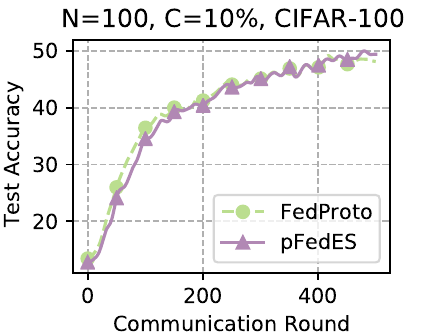}
\end{minipage}%
\vspace{-0.5em}
\caption{Average accuracy vs. communication rounds.}
\label{fig:compare-hetero-converge}
\vspace{-1em}
\end{figure}

\textbf{Average Accuracy}. The results in Tables~\ref{tab:compare-homo} and \ref{tab:compare-hetero} show that the average accuracy of all personalized heterogeneous local models in \methodname{} surpasses other baselines under both model-homogeneous and model-heterogeneous settings, by up to $1.00\%$ and $1.61\%$, respectively. Figure~\ref{fig:compare-hetero-converge} shows that \methodname{} converges to a higher average accuracy with faster convergence rates.

\begin{figure}[t]
\centering
\begin{minipage}[t]{0.5\linewidth}
\centering
\includegraphics[width=1.75in]{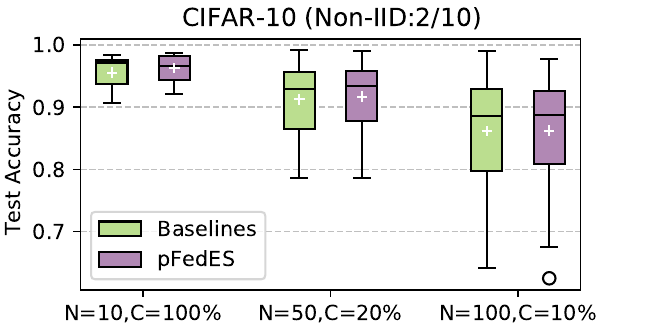}
\end{minipage}%
\begin{minipage}[t]{0.5\linewidth}
\centering
\includegraphics[width=1.75in]{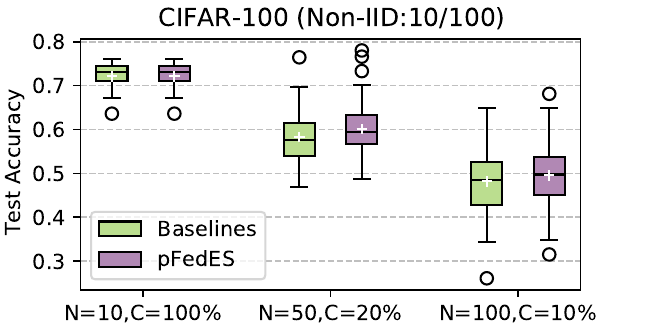}
\end{minipage}%
\vspace{-0.5em}
\caption{Accuracy distribution for individual clients.}
\label{fig:compare-hetero-individual}
\vspace{-1em}
\end{figure}

\textbf{Individual Accuracy}.
We use box plots to display the distribution of individual model accuracy in model-heterogeneous settings. As shown in Figure~\ref{fig:compare-hetero-individual}, `+' denotes the average accuracy of all clients for each algorithm. A small box length bounded by the upper quartile and the lower quartile indicates a more concentrated accuracy distribution across all clients with small variance. We observe that \methodname{} obtains a higher average accuracy and a lower variance than the optimal baselines ({\tt{Standalone}} or {\tt{FedProto}} in Table~\ref{tab:compare-hetero}) under most settings.

\begin{figure}[t]
\centering
\begin{minipage}[t]{0.5\linewidth}
\centering
\includegraphics[width=1.6in]{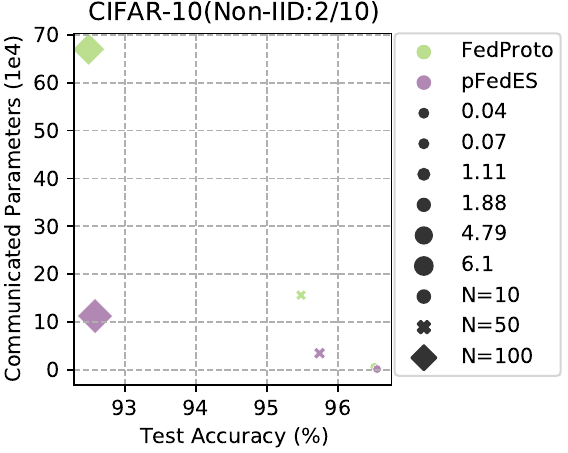}
\end{minipage}%
\begin{minipage}[t]{0.5\linewidth}
\centering
\includegraphics[width=1.6in]{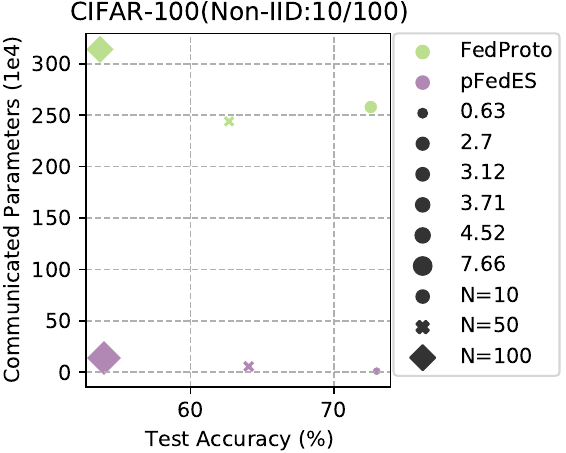}
\end{minipage}%
\vspace{-0.5em}
\caption{Trade-off among test accuracy, computational overhead and communication cost. The size of a marker reflects the computation FLOPs (1e9).}
\label{fig:compare-tradeoff}
\vspace{-1em}
\end{figure}

\textbf{Trade-off among Accuracy, Computation \& Communication Costs}. We compare \methodname{} and the state-of-the-art baseline {\tt{FedProto}} in terms of model accuracy, computational costs and communication costs. The target accuracy set for $N=\{10, 50, 100\}$ on CIFAR-10 dataset is $90\%$ and that set for $N=\{10, 50, 100\}$ on CIFAR-100 dataset are $\{70\%, 60\%, 50\%\}$. As shown in Figure~\ref{fig:compare-tradeoff}, \methodname{} consistently achieves the highest model accuracy with far lower communication costs than {\tt{FedProto}}, while incurring similar computation costs. This indicates that \methodname{} strikes the best trade-off among the three metrics. Compared with {\tt{FedProto}}, \methodname{} incurs only $\frac{1}{224}$ communication and $\frac{1}{5.85}$ computational costs (i.e., $99.6\%$ communication and $82.9\%$ computational cost savings), due to its faster convergence and lower per-round costs.

\begin{figure}[t]
\centering
\begin{minipage}[t]{0.5\linewidth}
\centering
\includegraphics[width=1.6in]{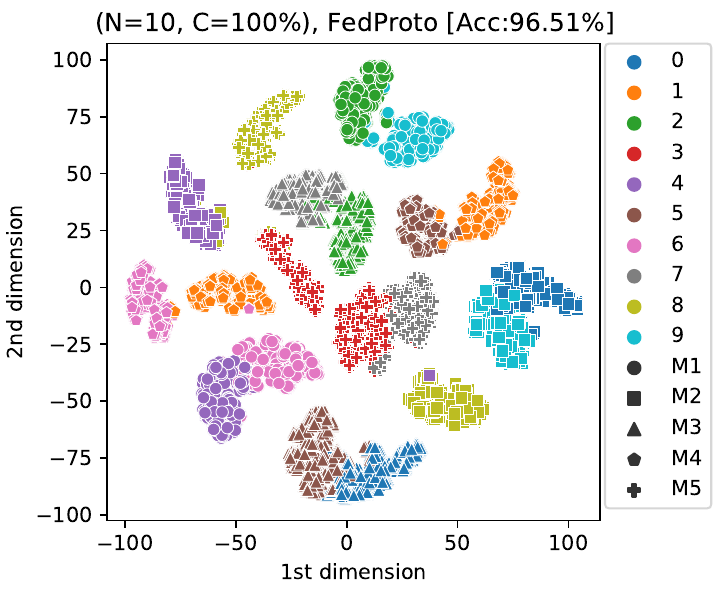}
\end{minipage}%
\begin{minipage}[t]{0.5\linewidth}
\centering
\includegraphics[width=1.6in]{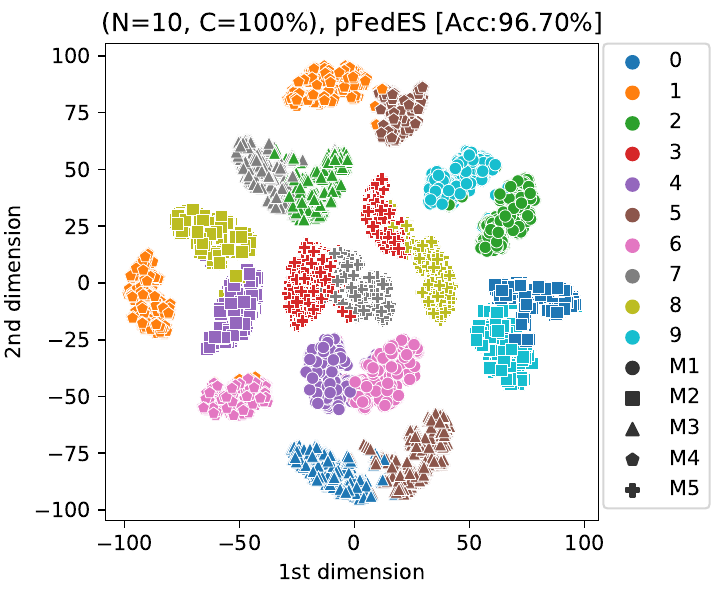}
\end{minipage}%
\caption{T-SNE representation visualization for {\tt{FedProto}} and \methodname{} on CIFAR-10 (Non-IID: 2/10).}
\label{fig:compare-TSNE}
\end{figure}

\textbf{Visualizing Personalization}.
In model-heterogeneous settings, we extract every sample representation from each FL client under \methodname{} and {\tt{FedProto}}. Then, we leverage \textit{T-SNE} \cite{TSNE-JMLR} to reduce the dimensionality of the extracted representations from $500$ to $2$, and visualize the results. Since CIFAR-100 includes 100 classes of samples, we focus on visualizing the results on CIFAR-10 (non-IID: 2/10) with $N=10$ in Figure~\ref{fig:compare-TSNE}. We observe that most clusters under the two methods consist of representations from a client's two seen classes of samples, indicating that each client's heterogeneous local model has strong \textit{personalization} capability. The two seen class representations within most clusters under the two methods satisfy ``intra-class compactness and inter-class separation'', reflecting that every client can classify its seen classes well. Generally, \methodname{} achieves better classification boundaries than {\tt{FedProto}}.

\begin{figure}[t]
\centering
\includegraphics[width=1\linewidth]{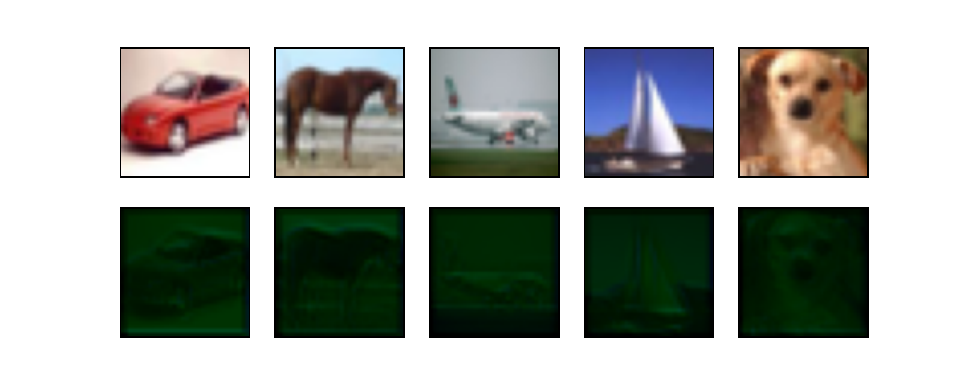}
\vspace{-0.5em}
\caption{Visualization of original data and enhanced data.}
\label{fig:visual-extractor}
\vspace{-1em}
\end{figure}

\textbf{Visualizing Enhanced Data}. We store the final global feature extractor trained on CIFAR-10 under the model-heterogeneous FL scenario with $N=100$. We sample $5$ original images randomly from CIFAR-10 and input them into the final global feature extractor to generate the corresponding enhanced data. The two lines in Figure~\ref{fig:visual-extractor} present the visualized results of the original data and the generated enhanced data. It can be observed that enhanced images embedded with global information still retain partial features of original images, indicating that the shared homogeneous feature extractor learns global and local personalized knowledge effectively. Besides, valid enhanced data increase the number of training samples, thereby boosting the accuracy of heterogeneous local models.

\subsection{Case Studies}
\subsubsection{Robustness to Non-IID Data}
\begin{figure}[t]
\centering
\begin{minipage}[t]{0.5\linewidth}
\centering
\includegraphics[width=\linewidth]{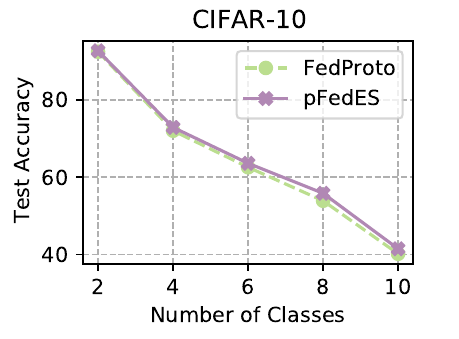}
\end{minipage}%
\begin{minipage}[t]{0.5\linewidth}
\centering
\includegraphics[width=\linewidth]{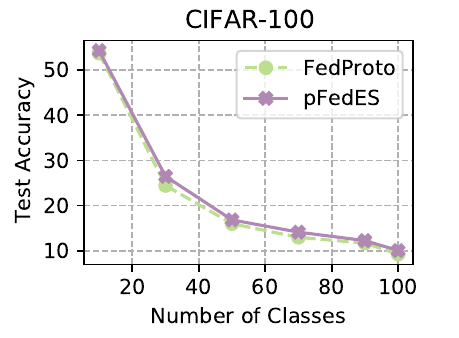}
\end{minipage}%
\vspace{-0.5em}
\caption{Robustness to Non-IID data.}
\label{fig:case-noniid}
\end{figure}

We evaluate the robustness of \methodname{} and {\tt{FedProto}} under different non-IID data settings with $(N=100, C=10\%)$. We vary the number of classes seen by each client as $\{2, 4, 6, 8, 10\}$ on CIFAR-10 and $\{10, 30, 50, 70, 90, 100\}$ on CIFAR-100. Figure~\ref{fig:case-noniid} shows that \methodname{} consistently outperforms {\tt{FedProto}}, demonstrating its robustness to non-IID data. As the non-IID degree decreases (i.e., the number of classes seen by each client increases), the accuracy drops since more IID data enhances generalization and reduces personalization.

\subsubsection{Robustness to Client Participant Rates}
\begin{figure}[t]
\centering
\begin{minipage}[t]{0.5\linewidth}
\centering
\includegraphics[width=\linewidth]{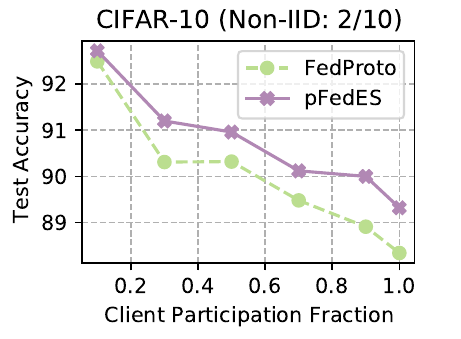}
\end{minipage}%
\begin{minipage}[t]{0.5\linewidth}
\centering
\includegraphics[width=\linewidth]{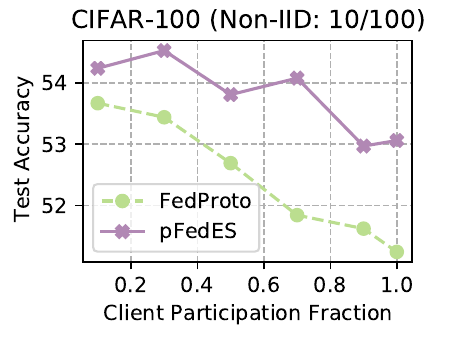}
\end{minipage}%
\vspace{-0.5em}
\caption{Robustness to client participation rates.}
\label{fig:case-frac}
\vspace{-0.5em}
\end{figure}

We also test the robustness of \methodname{} and {\tt{FedProto}} against different client participant rates $C$ under $(N=100)$ on CIFAR-10 (non-IID: 2/10) and CIFAR-100 (non-IID: 10/100). We vary the client participant rates as $C=\{0.1, 0.3, 0.5, 0.7, 0.9, 1\}$. Figure~\ref{fig:case-frac} shows that \methodname{} consistently outperforms {\tt{FedProto}}, especially on more complicated CIFAR-100, verifying its robustness to changes in client participant rates. Besides, as the client participant rates increase, model accuracy drops as more participating clients provide more IID local data, which also improves generalization and reduces personalization.

%% file: sec/7_conclusion.tex
\section{Conclusions and Future Work}
In this paper, we proposed a novel MHPFL approach, \methodname{}, based on sharing homogeneous feature extractors with efficient privacy preservation, and communication and computation cost savings. It enables each client to iteratively train a homogeneous feature extractor and heterogeneous local model to exchange global and local knowledge. Aggregating the homogeneous local feature extractors from clients fuses knowledge across their heterogeneous local models. Theoretical analysis proves its convergence. Extensive experiments demonstrate that \methodname{} obtains the highest model accuracy, while incurring the lowest communication and computation costs.

In subsequent research, we plan to extend \methodname{} in two ways: 1) exploring more efficient structures for the homogeneous feature extractor; and 2) optimizing the pattern of training the homogeneous feature extractor and the local heterogeneous model, in order to further improve model performance and reduce the additional computational costs incurred by training the homogeneous feature extractors.

%% file: sec/X_suppl.tex
\clearpage
\setcounter{page}{1}
\maketitlesupplementary

\section{Proof for Lemma~\ref{lemma:LocalTraining}} \label{sec:proof-lemma}

\begin{proof}
As formulated in Eq.~(\ref{eq:miu}), the local heterogeneous model of an arbitrary client $k$ is updated by
\begin{equation}\label{eq:gradient-descent}
    \omega_{t+1}=\omega_t-\eta g_{\omega}^t=\omega_t-\eta\nabla((1-\mu)\cdot\mathcal{L}_{\omega_t}+\mu\cdot\mathcal{L}_{\varphi_t}).
\end{equation}

Based on Assumption~\ref{assump:Lipschitz} and Eq.~(\ref{eq:gradient-descent}), we can get
\begin{equation}
\begin{aligned}
\mathcal{L}_{t E+1} & \leq \mathcal{L}_{t E+0}+\langle\nabla \mathcal{L}_{t E+0},(\omega_{t E+1}-\omega_{t E+0})\rangle+\frac{L_1}{2}\|\omega_{t E+1}-\omega_{t E+0}\|_2^2 \\
& =\mathcal{L}_{t E+0}-\eta\langle\nabla \mathcal{L}_{t E+0}, \nabla((1-\mu) \cdot \mathcal{L}_{\omega_{t E+0}}+\mu \cdot \mathcal{L}_{\varphi_{t E+0}})\rangle+\frac{L_1 \eta^2}{2}\|\nabla((1-\mu) \cdot \mathcal{L}_{\omega_{t E+0}}+\mu \cdot \mathcal{L}_{\varphi_{t E+0}})\|_2^2.
\end{aligned}
\end{equation}

Take the expectations of random variable $\xi_{tE+0}$ on both sides, we have
\begin{equation}
\small
\begin{aligned}
\mathbb{E}[\mathcal{L}_{t E+1}] & \leq \mathcal{L}_{t E+0}-\eta \mathbb{E}[\langle\nabla \mathcal{L}_{t E+0}, \nabla((1-\mu) \cdot \mathcal{L}_{\omega_{t E+0}}+\mu \cdot \mathcal{L}_{\varphi_{t E+0}})\rangle]+\frac{L_1 \eta^2}{2} \mathbb{E}[\|\nabla((1-\mu) \cdot \mathcal{L}_{\omega_{t E+0}}+\mu \cdot \mathcal{L}_{\varphi_{t E+0}})\|_2^2] \\
& \stackrel{(a)}{\leq} \mathcal{L}_{t E+0}-\eta \mathbb{E}[\langle\nabla \mathcal{L}_{t E+0}, \nabla((1-\mu) \cdot \mathcal{L}_{\omega_{t E+0}})\rangle]+\frac{L_1 \eta^2}{2} \mathbb{E}[\|\nabla((1-\mu) \cdot \mathcal{L}_{\omega_{t E+0}}+\mu \cdot \mathcal{L}_{\varphi_{t E+0}})\|_2^2] \\
& =\mathcal{L}_{t E+0}-\eta (1-\mu)\|\nabla \mathcal{L}_{\omega_{t E+0}}\|_2^2+\frac{L_1 \eta^2}{2} \mathbb{E}[\|\nabla((1-\mu) \cdot \mathcal{L}_{\omega_{t E+0}}+\mu \cdot \mathcal{L}_{\varphi_{t E+0}})\|_2^2] \\
& \stackrel{(b)}{=} \mathcal{L}_{t E+0}-\eta (1-\mu)\|\nabla \mathcal{L}_{\omega_{t E+0}}\|_2^2+\frac{L_1 \eta^2}{2}(Var(\nabla((1-\mu) \cdot \mathcal{L}_{\omega_{t E+0}}+\mu \cdot \mathcal{L}_{\varphi_{t E+0}}))+\|\nabla((1-\mu) \cdot \mathcal{L}_{\omega_{t E+0}}+\mu \cdot \mathcal{L}_{\varphi_{t E+0}})\|_2^2) \\
& \stackrel{(c)}{\leq} \mathcal{L}_{t E+0}-\eta (1-\mu)\|\nabla \mathcal{L}_{\omega_{t E+0}}\|_2^2+\frac{L_1 \eta^2}{2}((\sigma^2+\delta^2)+\|\nabla((1-\mu) \cdot \mathcal{L}_{\omega_{t E+0}}+\mu \cdot \mathcal{L}_{\varphi_{t E+0}})\|_2^2) \\
& \stackrel{(d)}{\leq} \mathcal{L}_{t E+0}-\eta (1-\mu)\|\nabla \mathcal{L}_{\omega_{t E+0}}\|_2^2+\frac{L_1 \eta^2}{2}((\sigma^2+\delta^2)+\|\nabla((1-\mu) \cdot \mathcal{L}_{\omega_{t E+0}})\|_2^2) \\
& =\mathcal{L}_{t E+0}+(\frac{L_1 \eta^2 (1-\mu)^2}{2}-\eta (1-\mu))\|\nabla \mathcal{L}_{\omega_{t E+0}}\|_2^2+\frac{L_1 \eta^2(\sigma^2+\delta^2)}{2} \\
& \stackrel{(e)}{=}\mathcal{L}_{t E+0}+(\frac{L_1 \eta^2 \tilde{\mu}^2}{2}-\eta \tilde{\mu})\|\nabla \mathcal{L}_{\omega_{t E+0}}\|_2^2+\frac{L_1 \eta^2(\sigma^2+\delta^2)}{2}, 
\end{aligned}
\end{equation}
where $(a):\mu\in(0,0.5]$, $(1-\mu)\in[0.5,1)$, $\ell_{\varphi_{tE+0}}\geq0$, so $\mu\cdot\ell_{\varphi_{tE+0}}\geq0$. $(b)$ follows from $Var(x)=\mathbb{E}[x^2]-(\mathbb{E}[x]^2)$. $(c)$ follows from Assumption~\ref{assump:Unbiased}. $(d)$ are the same as $(a)$. $(e)$: we denote $1-\mu=\tilde{\mu}, \tilde{\mu}\in[0.5,1)$.

Take the expectations of the heterogeneous local model $\omega$ on both sides across $E$ local iterations, we have
\begin{equation}\label{eq:lemma}
\mathbb{E}[\mathcal{L}_{(t+1) E}] \leq \mathcal{L}_{t E+0}+(\frac{L_1 \eta^2 \tilde{\mu}^2}{2}-\eta \tilde{\mu}) \sum_{e=0}^{E-1}\|\nabla \mathcal{L}_{t E+e}\|_2^2+\frac{L_1 \eta^2(\sigma^2+\delta^2)}{2}.
\end{equation}
\end{proof}

\section{Proof for Theorem~\ref{theorem:non-convex}}\label{sec:proof-theorem}
\begin{proof}
Eq.~(\ref{eq:lemma}) can be adjusted further as
\begin{equation}
\sum_{e=0}^{E-1}\|\nabla \mathcal{L}_{t E+e}\|_2^2 \leq \frac{\mathcal{L}_{t E+0}-\mathbb{E}[\mathcal{L}_{(t+1) E}]+\frac{L_1 \eta^2(\sigma^2+\delta^2)}{2}}{\eta \tilde{\mu}-\frac{L_1 \eta^2 \tilde{\mu}^2}{2}}.
\end{equation}

Take the expectations of the heterogeneous local model $\omega$ on both sides across $T$ communication rounds, we have
\begin{equation}
\frac{1}{T} \sum_{t=0}^{T-1} \sum_{e=0}^{E-1}\|\nabla \mathcal{L}_{t E+e}\|_2^2 \leq \frac{\frac{1}{T} \sum_{t=0}^{T-1}(\mathcal{L}_{t E+0}-\mathbb{E}[\mathcal{L}_{(t+1) E}])+\frac{L_1 \eta^2(\sigma^2+\delta^2)}{2}}{\eta \tilde{\mu}-\frac{L_1 \eta^2 \tilde{\mu}^2}{2}}.
\end{equation}

Let $\Delta=\mathcal{L}_{t=0} - \mathcal{L}^* > 0$, then $\sum_{t=0}^{T-1}(\mathcal{L}_{t E+0}-\mathbb{E}[\mathcal{L}_{(t+1) E}]) \leq \Delta$, so we have
\begin{equation}\label{eq:T}
\frac{1}{T} \sum_{t=0}^{T-1} \sum_{e=0}^{E-1}\|\nabla \mathcal{L}_{t E+e}\|_2^2 \leq \frac{\frac{\Delta}{T}+\frac{L_1 \eta^2(\sigma^2+\delta^2)}{2}}{\eta \tilde{\mu}-\frac{L_1 \eta^2 \tilde{\mu}^2}{2}}.
\end{equation}

If the above equation can converge to a constant $\epsilon$, i.e.,
\begin{equation}
\frac{1}{T} \sum_{t=0}^{T-1} \sum_{e=0}^{E-1}\|\nabla \mathcal{L}_{t E+e}\|_2^2 \leq \frac{\frac{\Delta}{T}+\frac{L_1 \eta^2(\sigma^2+\delta^2)}{2}}{\eta \tilde{\mu}-\frac{L_1 \eta^2 \tilde{\mu}^2}{2}}<\epsilon, 
\end{equation}
then
\begin{equation}
T>\frac{2 \Delta}{\eta \tilde{\mu}(2-L_1 \eta \tilde{\mu})(\epsilon-\frac{\eta L_1(\sigma^2+\delta^2)}{\tilde{\mu}(2-L_1 \eta \tilde{\mu})})}.
\end{equation}

Since $T>0, \Delta>0$, so we get
\begin{equation}
\eta \tilde{\mu}(2-L_1 \eta \tilde{\mu})(\epsilon-\frac{\eta L_1(\sigma^2+\delta^2)}{\tilde{\mu}(2-L_1 \eta \tilde{\mu})})>0.
\end{equation}

After solving the above inequality, we can get
\begin{equation}
\eta<\frac{2 \epsilon \tilde{\mu}}{L_1(\sigma^2+\delta^2+\tilde{\mu}^2 \epsilon)}.
\end{equation}

Since $\epsilon,\tilde{\mu},\ L_1,\ \sigma^2,\ \delta^2 > 0$ are both constants, the learning rate $\eta$ of the local heterogeneous model has solutions.

Therefore, when the learning rate of the local heterogeneous model satisfies the above condition, an arbitrary client's local heterogeneous local can converge. In addition, on the right side of Eq.~(\ref{eq:T}), except for $\frac{\Delta}{T}$, $\Delta$ and other items are both constants, so the non-convex convergence rate $\epsilon \sim \mathcal{O}(\frac{1}{T})$.
\end{proof}

\section{Sensitivity to Hyperparameters}\label{app:hyperparameter}
\begin{figure}[h]
\centering
\begin{minipage}[t]{0.25\linewidth}
\centering
\includegraphics[width=1.8in]{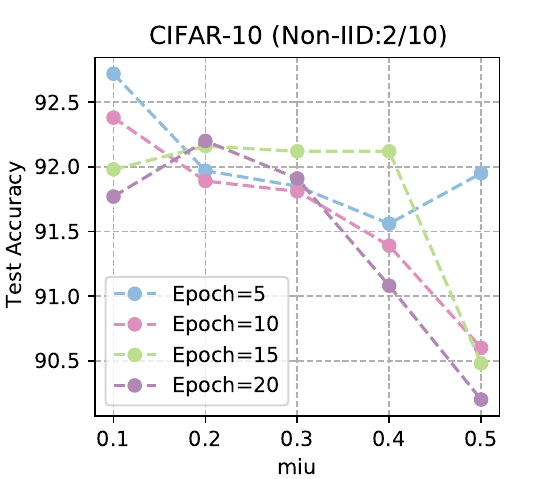}
\end{minipage}%
\begin{minipage}[t]{0.25\linewidth}
\centering
\includegraphics[width=1.8in]{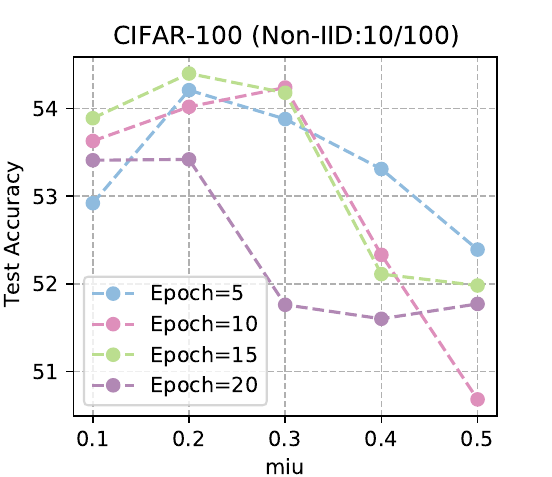}
\end{minipage}%
\begin{minipage}[t]{0.25\linewidth}
\centering
\includegraphics[width=1.8in]{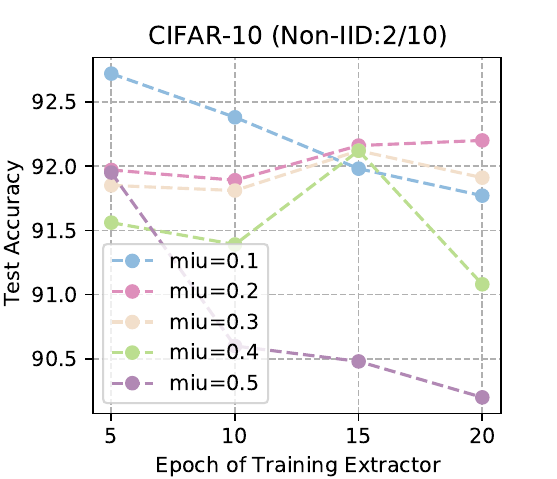}
\end{minipage}%
\begin{minipage}[t]{0.25\linewidth}
\centering
\includegraphics[width=1.8in]{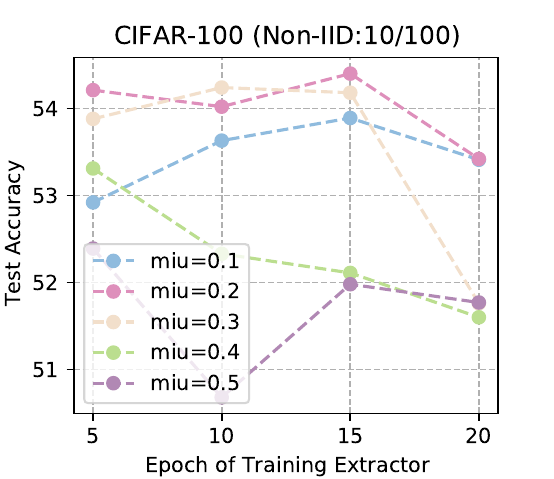}
\end{minipage}%
\caption{Sensitivity to hyperparameters $\mu$ and $E_{fe}$.}
\label{fig:case-sensitivity}
\end{figure}

\methodname{} involves two key hyperparameters: 1) $\mu$, the \textit{weight of the loss} for the combination of the frozen feature extractor and the training local heterogeneous model in step \ding{192} of iterative training; 2) $E_{fe}$, the \textit{training epochs} for the combination of the training feature extractor and the frozen local heterogeneous model in step \ding{193} of iterative training.
Take the model-heterogeneous experiments with $(N=100,C=10\%)$ as examples, Fig.~\ref{fig:case-sensitivity} displays that the average test accuracy of \methodname{} varies as $\mu=\{0.1,0.2,0.3,0.4,0.5\}$ and $E_{fe}=\{5,10,15,20\}$. Model accuracy decreases as $\mu$ rises since a larger $\mu$ emphasizes more globally generalized knowledge and weakens local personalized knowledge. Model accuracy also drops as $E_{fe}$ increases since a larger $E_{fe}$ may lead the combination of the training feature extractor and the frozen local heterogeneous model to be overfitting.

